\documentclass{article}

\usepackage{arydshln}
\usepackage{subcaption}
\usepackage{amsmath}
\usepackage{listings}
\usepackage{graphicx}

\PassOptionsToPackage{square,sort,comma,numbers}{natbib}

     \usepackage[final]{neurips_2019}

\usepackage[utf8]{inputenc} 
\usepackage[T1]{fontenc}    
\usepackage{hyperref}       
\usepackage{url}            
\usepackage{booktabs}       
\usepackage{amsfonts}       
\usepackage{nicefrac}       
\usepackage{microtype}      
\usepackage[table,xcdraw]{xcolor}
\usepackage{multirow}
\usepackage{xcolor}

\usepackage{hhline}

\usepackage{array, makecell, cellspace}
\usepackage{array}

\makeatletter
\newcommand{\thickhline}{%
    \noalign {\ifnum 0=`}\fi \hrule height 1pt
    \futurelet \reserved@a \@xhline
}
\newcolumntype{"}{@{\hskip\tabcolsep\vrule width 1pt\hskip\tabcolsep}}
\makeatother

\newcommand{\cmmnt}[1]{}

\newcommand{\sigmoid}{\ensuremath{\mathrm{sigm}}}
\renewcommand{\tanh}{\ensuremath{\mathrm{tanh}}}
\renewcommand{\exp}{\ensuremath{\mathrm{exp}}}

\newcommand{\identity}{\ensuremath{\mathrm{identity}}}

\newcommand{\relu}{\ensuremath{\mathrm{ReLU}}}

\newcommand{\safebiggestsavingpercent}{\ensuremath{37\%}}
\newcommand{\safesmallestsavingpercent}{\ensuremath{10\%}}
\newcommand{\rangedbiggestsavingpercent}{\ensuremath{53\%}}
\newcommand{\rangedsmallestsavingpercent}{\ensuremath{20\%}}

\newcommand{\approxfn}{\begingroup
  \catcode`_=12 \doapproxfn}
\newcommand{\doapproxfn}[1]{%
  \texttt{#1}%
  \endgroup
}

\title{Approximating Activation Functions}

\newcounter{@inst}
\newcounter{@auth}

\newdimen\instindent
\newbox\authrun
\newtoks\authorrunning
\newtoks\tocauthor
\newbox\titrun
\newtoks\titlerunning
\newtoks\toctitle

\def\clearheadinfo{\gdef\@author{No Author Given}%
                   \gdef\@title{No Title Given}%
                   \gdef\@subtitle{}%
                   \gdef\@institute{No Institute Given}%
                   \gdef\@thanks{}%
                   \global\titlerunning={}\global\authorrunning={}%
                   \global\toctitle={}\global\tocauthor={}}
                   
                   \def\institute#1{\gdef\@institute{#1}}

\def\institutename{\par
 \begingroup
 \parskip=\z@
 \parindent=\z@
 \setcounter{@inst}{1}%
 \def\and{\par\stepcounter{@inst}%
 \noindent$^{\the@inst}$\enspace\ignorespaces}%
 \setbox0=\vbox{\def\thanks##1{}\@institute}%
 \ifnum\c@@inst=1\relax
   \gdef\fnnstart{0}%
 \else
   \xdef\fnnstart{\c@@inst}%
   \setcounter{@inst}{1}%
   \noindent$^{\the@inst}$\enspace
 \fi
 \ignorespaces
 \@institute\par
 \endgroup}

\def\@fnsymbol#1{\ensuremath{\ifcase#1\or\star\or{\star\star}\or
   {\star\star\star}\or \dagger\or \ddagger\or
   \mathchar "278\or \mathchar "27B\or \|\or **\or \dagger\dagger
   \or \ddagger\ddagger \else\@ctrerr\fi}}

\def\email#1{{\tt#1}}

\author{Nicholas Gerard Timmons \and
Andrew Rice}
\institute{Dept. of Computer Science and Technology, University of Cambridge, Cambridge, UK
\email{ngt26@cam.ac.uk}\\
\url{https://www.cl.cam.ac.uk/}}

\begin{document}
\maketitle

\begin{abstract}
ReLU is widely seen as the default choice for activation functions in neural networks. However, there are cases where more complicated functions are required. In particular, recurrent neural networks (such as LSTMs) make extensive use of both hyperbolic tangent and sigmoid functions. These functions are expensive to compute. We used function approximation techniques to develop replacements for these functions and evaluated them empirically on three popular network configurations. We find safe approximations that yield a \safesmallestsavingpercent{} to \safebiggestsavingpercent{} improvement in training times on the CPU. These approximations were suitable for all cases we considered and we believe are appropriate replacements for all networks using these activation functions. We also develop ranged approximations which only apply in some cases due to restrictions on their input domain. Our ranged approximations yield a performance improvement of \rangedsmallestsavingpercent{} to \rangedbiggestsavingpercent{} in network training time. Our functions also match or considerably out-perform the ad-hoc approximations used in Theano and the implementation of Word2Vec.
\end{abstract}

\section{Introduction}
Function approximation is a software engineering technique in which we seek to replace complex functions with faster, but less accurate, alternatives. For example, consider  $f(x) = \sin{x}$. This is a challenging and complex function to implement correctly and much research has been published on algorithms for achieving minimal error with correct rounding~\cite{de2005towards,de2004fast,defour2004proposal}. However, if the caller were able to tolerate some error we could replace the implementation of $f(x)$ with some terms from the Taylor Series for $\sin{x}$. This yields faster execution at the cost of some loss of accuracy.

Much of the research on function approximation incorporates a significant amount of mathematical reasoning to argue that the approximation falls within a worst-case error bound in all cases: the Taylor Series example above would not be used in practice because its worst-case bound is too high.

However, there are many applications where such strict requirements are unnecessary and in this case one can make use of simpler (and considerably faster) approximations that are `good enough'~\cite{sampson2013good}. In this paper we consider one such application: that of activation functions in neural networks. Not only is a neural network inherently tolerant to error but practitioners will admit that there is a certain art to the selection of activation functions (and other hyper-parameters) and thus a certain leeway in their accuracy.

Function approximation has been applied to activation functions before. The implementation of Word2Vec makes use of a lookup table for approximating the exponential function and the popular machine learning library Theano includes an approximation of the \sigmoid{} function called \approxfn{ultra_fast_sigmoid}. Google also released some research very recently where they showed a piecewise approximations of their `Swish' activation function for use in their edge computing TensorFlow interface which shows the viability of some approximation for limited hardware~\cite{howard2019searching}. However, this paper is the first detailed study on the subject in this area, and shows that approximations can be used as drop-in replacements for current models. The approximations we present outperform all of the mentioned alternative approximation approaches.

In this paper we consider a range of approximations with a trade-off between complexity and accuracy. We study the overall impact on training and prediction time for popular neural network architectures. We identify two approximation options for each: a \emph{safe} approximation that works in all cases and a \emph{ranged} approximation that requires some restrictions to it's input domain. For all networks tested we find that our safe approximations outperform the standard implementations with an \safesmallestsavingpercent{} to \safebiggestsavingpercent{} improvement in training times. Where appropriate our ranged implementations provide an improvement of \rangedsmallestsavingpercent{} to \rangedbiggestsavingpercent{}. Our training and inference benchmarks are initially run on CPU as a proof of concept.

We believe that our \emph{safe} approximations are of particular relevance to library developers since they are better-performing drop-in replacements for existing functions. There is also a growing range of specialised hardware designed for accelerating machine learning tasks and in future work we would like to consider how well our approximations might translate into hardware implementations.

We provide an open-source implementation of our approach in Julia using the Flux\footnote{\url{https://fluxml.ai/}} machine learning library. We argue that these approximations are applicable to other libraries too and we include a brief study using TensorFlow~\cite{abadi2016tensorflow} as evidence. 

\section{Activation functions}  

Activation functions are used in neural networks to introduce non-linearity. For activation functions with bounded results, the popular choices are logistic, trigonometric and clamping functions with the sigmoid (\sigmoid{}) and hyperbolic tangent (\tanh{}) functions being used most frequently. For unbounded or partially bounded activations, the \relu{} function and it's variants are most popular. 

\relu{} is the dominant choice for an activation function but \sigmoid{} and \tanh{} still have significant use as gating functions in feed-forward and recurrent networks. 

We focus on \sigmoid{} and \tanh{} in particular as major components of LSTMs gating behaviour which in it's original form uses three calls to \sigmoid{} and two calls to \tanh{}~\cite{graves2013generating}.




For cases (such as LSTMs) where \relu{} is not appropriate the performance impact can be notable: in micro-benchmarks measuring function execution time we found \sigmoid{} and \tanh{} to be 3.6x and 7.9x more expensive than \relu{}. In this paper we focus on these two functions given their popularity and relatively high cost. 

A significant proportion of CPU time is spent in the computation of values in activation functions. For example, for a densely connected network with 51,000 trainable parameters using a \sigmoid{} activation function with two equal-sized hidden layers we measure that approximately 29\% of time is spent computing the \sigmoid{} function. For context, in the paper which introduced sequence-to-sequence learning for neural networks the authors use a 5-layer LSTM network which has 380 million trainable parameters~\cite{sutskever2014sequence}.

\section{Function Approximation}

A conventional implementation of a mathematical function will aim to compute results which arise from rounding actual values to the precision of the data-type used. For example, implementations of the C Math library often come with documentation stating to how many units-in-last-place the implementation of a specific function will return in the worst case. An application might be able to tolerate considerably more error than this but it can be difficult to prove conclusively. In this paper we seek to validate our approximations empirically. We believe this is appropriate since the `correct' choice of activation function is most often demonstrated empirically too. 

In addition to tolerating error in the function's outputs an application might also restrict the function's input domain. This provides further opportunity for approximation since it is only necessary to mimic the original function within the domain of interest.\cmmnt{Some neural networks have this property and some do not.} We therefore develop two classes of approximation. Our \emph{ranged} approximation produces higher performance under the assumption that the input domain is approximately -5.5 to 5.5. Our \emph{safe} approximation makes no assumptions about input domain and so is a drop-in replacement for all networks.

We consider two different approximation approaches: 1) taking a mathematically derived approximation of the function and then using the parameters to produce a range of alternative versions at varying precision; 2) performing a numerical regression to fit the given function. The mathematically defined approach produces more stable results but can be more complex. The regression approach produces often faster and simpler functions but the output result is heavily determined by the input parameters and function that is being replaced.

\subsection{\relu{}} We first consider an optimisation of \relu{} which transforms a max operator (which may use a branch) to a sum and a division. 
\begin{equation*}
    {(x + |x|)}*{0.5}
\end{equation*}
This may result in changing the return value from $x$ to a value near to $x$ under the rules of IEEE floating-point arithmetic~\cite{1985--ieee754}, but it also avoids the branch in the if-statement if the \approxfn{max} function is not properly implemented.
In our tests this resulted in an insignificant change and so we use the standard implementation of \relu{} as our baseline.

\subsection{\tanh{}}

\begin{figure}
    \centering
    \includegraphics[width=\textwidth]{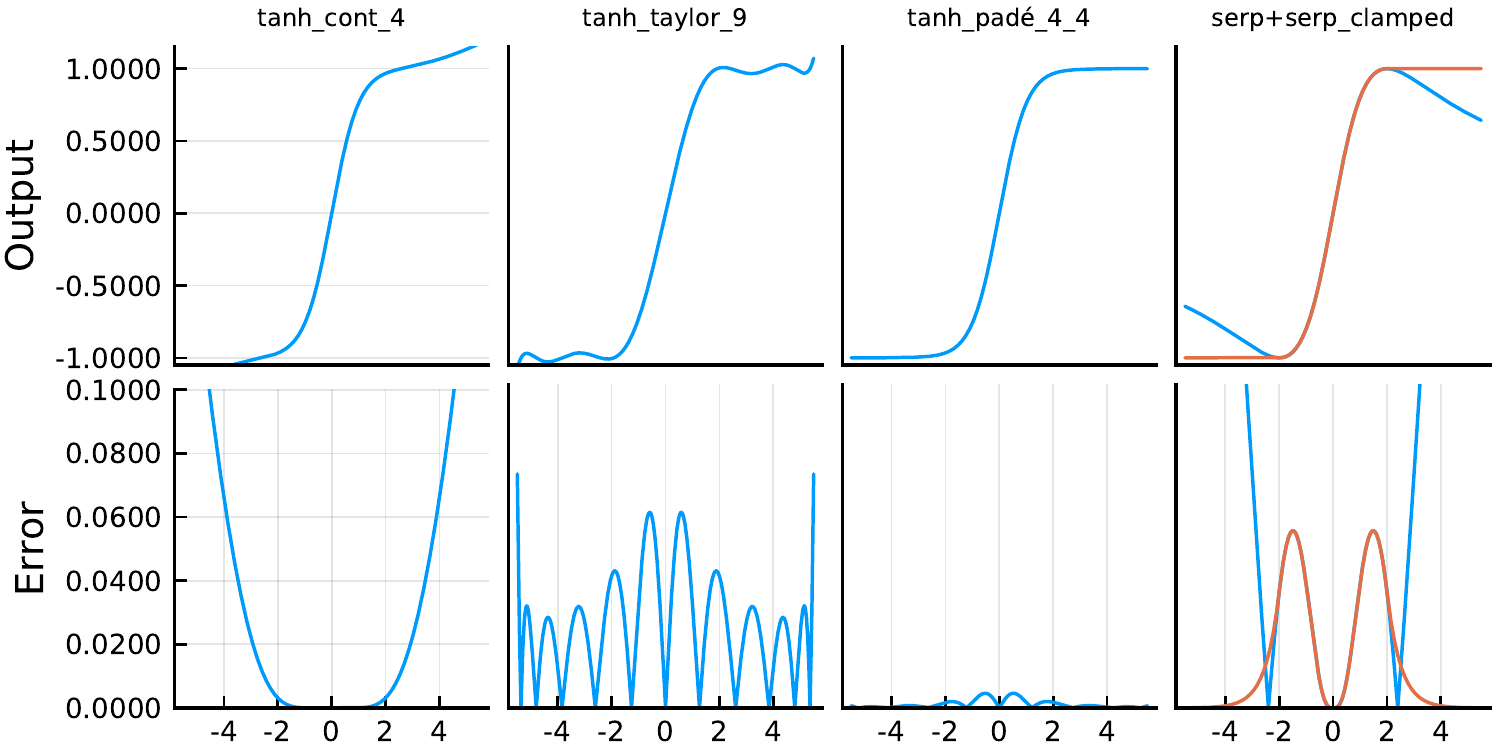}
    \caption{Approximation functions for \tanh{} and their associated error calculated as the absolute difference between the original \tanh{} function per input $x$.}
    \label{fig:tanh_approx}
\end{figure}

We consider four forms of approximation of \tanh{} which we show in Figure \ref{fig:tanh_approx} and describe below.

Firstly we implemented \tanh{} as Lambert's continued fraction. We limit ourselves to a finite number ($n$) of iterations depending on the precision we want and then simplify using a symbolic algebra package. We call this approximation \approxfn{tanh_cont_n}.
\begin{equation*}
  \frac{x}{1+\frac{x^2}{3+\frac{x^2}{5+\dots}}}
\end{equation*}
We also consider two forms of polynomial approximation of \tanh{}: 1) a Taylor Series of the form $\sum_{i=0}^{n} a_ix^i$ and; 2) Pad\'e approximants of the form $\frac{\sum_{i=0}^{n} a_ix^i}{\sum_{i=0}^{m} b_ix^i}$. We select a desired number of terms ($n,m$), choose a uniform sample of 5000 values in the range $-5.5$ to $5.5$ and then determine the values of the coefficients ($a_i,b_i$) using a least-squares fitting procedure. We choose this range based on the shape of the \tanh{} (and \sigmoid{}) which are very close to $-1$ and $1$ by this point. Using these two techniques yields the approximations \approxfn{tanh_taylor_n} and \approxfn{tanh_pade_n_m}. Minor variations in the number of points sampled and the range covered did not have a big impact on our results.

As a final option for \tanh{} we consider the Serpentine function (\approxfn{serp}) defined as
\begin{equation*}
    y = \frac{2x}{x^{2}+4}
\end{equation*}

As the \approxfn{serp} function does not fit \tanh{} beyond the main gradient, we also introduced a variant \approxfn{serp_clamp} which is clamped to the values $-1$ or $1$ for inputs outside of the range $-2$ to $2$.

%

\subsection{\sigmoid{}}

\begin{figure}
    \centering
    \includegraphics[width=0.75\textwidth]{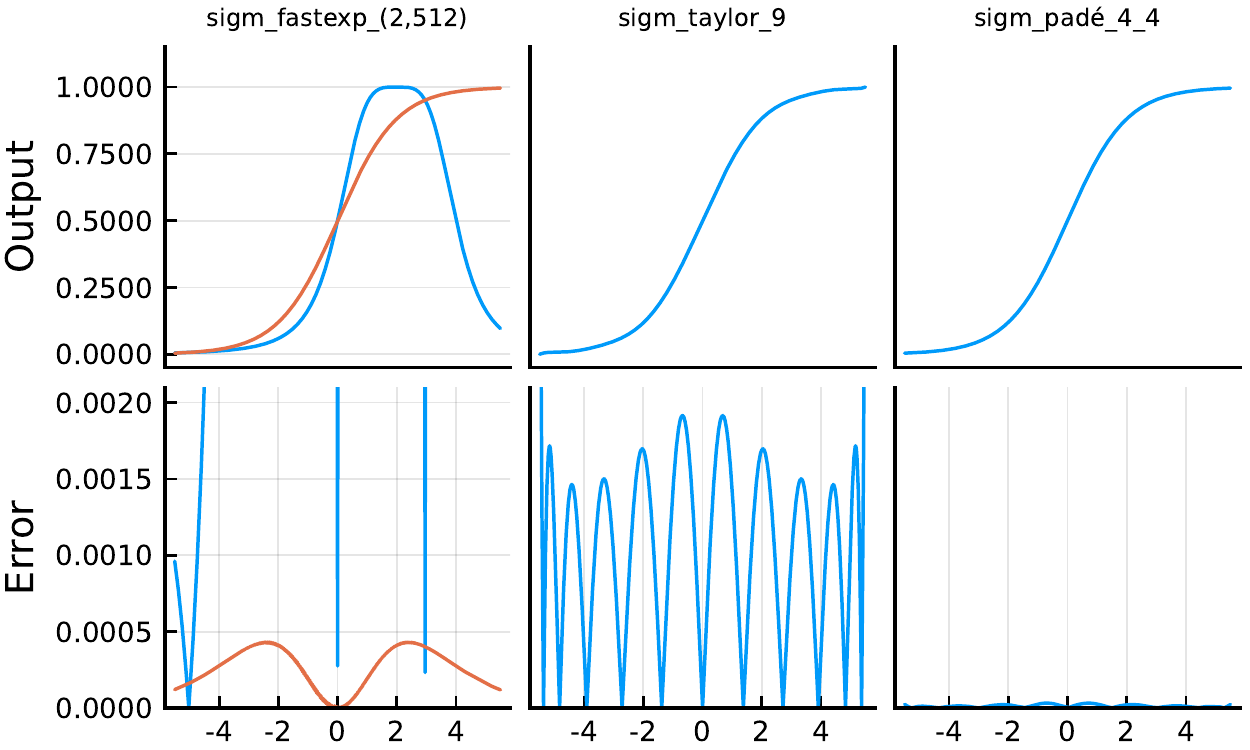}
    \caption{Approximation functions for \sigmoid{} and their associated error calculated as the absolute difference between the original \sigmoid{} function per input $x$.}
    \label{fig:SigmiodApproximations}
\end{figure}

Figure \ref{fig:SigmiodApproximations} shows three approximation forms for the \sigmoid{} function. These arise from the approximation of the \exp{} function included within the \sigmoid{} implementation. Here we show a well-known expansion of the \exp{} implementation for an infinite limit. %
We call approximations of this form \approxfn{sigmoid_fastexp_n}.
\begin{equation*}
\sigmoid(x) = \frac{1}{1+\exp(-x)} 
\mspace{70mu}
\exp(x) = \prod_{1}^{\lg(n)}\big(1+\frac{x}{n}\big)^n \\
\end{equation*}

We also generate fits for Talyor and Pad\'e polynomials in the same manner as for \tanh{}. This yields approximations \approxfn{sigmoid_taylor_n} and \approxfn{sigmoid_pade_n_m}.

We note that there is an alternative approach to the fast computation of the exponential function based on exploiting the definition of IEEE Floating-Point numbers~\cite{schraudolph1999fast}. This is no longer as effective as it once was due to it's reliance on a union structure to treat the floating-point number as both an integer for some parts and a floating-point number for others. This technique is not easily transferable to SIMD~\cite{nassimi1981data} where arrays of numbers are worked on in parallel and it is not always trivial to cast between different type representations in memory.

\section{Performance results\label{sec:PerformanceResults}}
In practice it is the performance of the whole network which is important rather than a particular activation function. For example, when training a neural network the ideal choice of activation function will result in the lowest loss for the least training time. This means that when selecting an approximation we are looking for a trade-off between the computation cost and the resulting error. Fast approximations with high error might cause a network to take longer to converge than slower approximations with a lower error. 

It is currently not possible to analytically determine the best choice of activation function for some network architecture. Similarly we are unable to prove that a particular approximation is a better choice in all cases. Instead we seek to empirically justify our choices through end-to-end measurements on three popular representative machine learning tasks covering three popular network architectures.

\textbf{MNIST Classifier} We consider the task of classifying images in the MNIST dataset and use a network comprising convolutional and dense layers inspired by the design of successful neural network structures~\cite{simard2003best} based on the implementation from a selection of provided models for Flux~\cite{innes2018flux}. While ReLU is sometimes used for convolutional image classification tasks for it's speed and simplicity, \tanh{} and \sigmoid{} have been used more commonly in the past~\cite{kalchbrenner2014convolutional, lawrence1997face, krizhevsky2012imagenet} and have some desirable properties which can reduce training times in some scenarios~\cite{ciresan2011flexible}. 

\textbf{MNIST Autoencoder} Autoencoders are common in the area of generative machine learning and are often evaluated using the MNIST dataset~\cite{lecun1998mnist}. We implemented a simple autoencoder to work with this dataset. Autoencoders are compatible with many different activations when structured with different configurations~\cite{szegedy2013intriguing, burda2015importance, ng2011sparse, vincent2008extracting, toderici2015variable, chen2012marginalized}. Our model is the simplest example of an autoencoder and as such is compatible with \relu{}, \sigmoid{} and \tanh{} activations.
        
\textbf{CharRNN} Sequence-to-sequence problems are another common task. We selected a common LSTM network layout with 2 hidden LSTM layers for text generation that takes a text dataset and learns to generate similar text. LSTM cells make use of both \sigmoid{} and \tanh{} activation functions and so we considered approximations to both.


\begin{table}
\begin{tabular}{ll|l|l|r|r|r|r|c|}
\cline{3-9}
                                                                                      & & \multicolumn{2}{c|}{\textbf{}}          & \multicolumn{2}{c|}{\textbf{Loss}} & \multicolumn{2}{c|}{\textbf{Time (s)}} & \\ \cline{2-9} 
\multicolumn{1}{l|}{}                                                                 & & \multicolumn{2}{c|}{\textbf{Activation}}              & Abs.   & Rel.   & Abs.     & Rel. & Choice            \\ \thickhline
\multicolumn{1}{|l|}{\multirow{14}{*}{\rotatebox[origin=c]{90}{\textbf{Convnet}}}}    & & \multicolumn{2}{l|}{ReLU             }                & 25.99  & 1.000   & 486.9 & 1.000  &                   \\ \cline{2-9} 
\multicolumn{1}{|l|}{}                                                                &\multicolumn{1}{l|}{\multirow{7}{*}{\rotatebox[origin=c]{90}{\textbf{Sigmoid}}}}  & 
                                                                                          \multicolumn{2}{l|}{sigm               }              & 20.23  & 1.000   & 889.5 & 1.000  &                   \\ \cdashline{3-9}
\multicolumn{1}{|l|}{}                                                                & & \multicolumn{2}{l|}{\textbf{sigm\_fastexp\_2}   }     & 17.21  & 0.851   & 643.5 & 0.723  & \textbf{Ranged}   \\ \cdashline{3-9} 
\multicolumn{1}{|l|}{}                                                                & & \multicolumn{2}{l|}{\textbf{sigm\_fastexp\_512} }     & 20.39  & 1.008   & 798.5 & 0.898  & \textbf{Safe}     \\ \cdashline{3-9} 
\multicolumn{1}{|l|}{}                                                                & & \multicolumn{2}{l|}{sigm\_taylor\_9    }              &$\infty$& $\infty$& -     & -      &                   \\ \cdashline{3-9} 
\multicolumn{1}{|l|}{}                                                                & & \multicolumn{2}{l|}{sigm\_pade\_4\_4   }              & 21.00  & 1.038   & 702.9 & 0.790  &                   \\ \cdashline{3-9} 
\multicolumn{1}{|l|}{}                                                                & & \multicolumn{2}{l|}{ultra\_fast\_sigmoid$\dagger$}    & 20.63  & 1.020   & 743.2 & 0.836  &                   \\ \cdashline{3-9}
\multicolumn{1}{|l|}{}                                                                & & \multicolumn{2}{l|}{word2vec$\dagger$}                & 456.1  & 22.55   & 833.9 & 0.937  &                   \\ \cline{2-9}
\multicolumn{1}{|l|}{}                                                                &\multicolumn{1}{l|}{\multirow{6}{*}{\rotatebox[origin=c]{90}{\textbf{Tanh}}}} &
                                                                                          \multicolumn{2}{l|}{tanh             }                & 16.64  & 1.000   & 1126  & 1.000  &                   \\ \cdashline{3-9}[6pt/3pt]  
\multicolumn{1}{|l|}{}                                                                & & \multicolumn{2}{l|}{tanh\_cont\_4    }                & 16.54  & 0.994   & 654.0 & 0.581  &                   \\ \cdashline{3-9}[6pt/3pt]  
\multicolumn{1}{|l|}{}                                                                & & \multicolumn{2}{l|}{tanh\_taylor\_9  }                &$\infty$&$\infty$ & -     & -      &                   \\ \cdashline{3-9}[6pt/3pt]  
\multicolumn{1}{|l|}{}                                                                & & \multicolumn{2}{l|}{\textbf{tanh\_pade\_4\_4} }       & 13.98  & 0.840   & 712.8 & 0.633  & \textbf{Safe}     \\ \cdashline{3-9}[6pt/3pt]  
\multicolumn{1}{|l|}{}                                                                & & \multicolumn{2}{l|}{\textbf{serp}       }             & 14.93  & 0.897   & 523.7 & 0.465  & \textbf{Ranged}   \\ \cdashline{3-9}[6pt/3pt]  
\multicolumn{1}{|l|}{}                                                                & & \multicolumn{2}{l|}{serp\_clamp}                      & 18.89  & 1.135   & 604.8 & 0.537  &                   \\ \thickhline

\multicolumn{1}{|l|}{\multirow{14}{*}{\rotatebox[origin=c]{90}{\textbf{Autoencoder}}}}& & \multicolumn{2}{l|}{ReLU}                             &  1.441 & 1.000 & 25.87  & 1.000 & - \\ \cline{2-9} 
\multicolumn{1}{|l|}{}                                                                & \multicolumn{1}{l|}{\multirow{7}{*}{\rotatebox[origin=c]{90}{\textbf{Sigmoid}}}} & 
                                                                                          \multicolumn{2}{l|}{sigm               }              &  4.166 & 1.000 & 35.54  & 1.000      &                   \\ \cdashline{3-9}[6pt/3pt]  
\multicolumn{1}{|l|}{}                                                                & & \multicolumn{2}{l|}{\textbf{sigm\_fastexp\_2}   }     &  3.924 & 1.006 & 28.33  & 0.797         & \textbf{Ranged}\\ \cdashline{3-9}[6pt/3pt]  
\multicolumn{1}{|l|}{}                                                                & & \multicolumn{2}{l|}{\textbf{sigm\_fastexp\_512} }     &  4.167 & 1.000 & 31.47  & 0.886         & \textbf{Safe}  \\ \cdashline{3-9}[6pt/3pt]  
\multicolumn{1}{|l|}{}                                                                & & \multicolumn{2}{l|}{sigm\_taylor\_9    }              &  6.788 & 1.630 & 32.74  & 0.921         &                \\ \cdashline{3-9}[6pt/3pt]  
\multicolumn{1}{|l|}{}                                                                & & \multicolumn{2}{l|}{sigm\_pade\_4\_4   }              &  4.161 & 0.999 & 30.56  & 0.860         &                \\ \cdashline{3-9}[6pt/3pt] 
\multicolumn{1}{|l|}{}                                                                & & \multicolumn{2}{l|}{ultra\_fast\_sigmoid$\dagger$}    &  4.210 & 1.011 & 32.48  & 0.914      &                   \\ \cdashline{3-9}[6pt/3pt] 
\multicolumn{1}{|l|}{}                                                                & & \multicolumn{2}{l|}{word2vec$\dagger$}                &  13.94 & 3.347 & 32.58  & 0.917      &                   \\ \cline{2-9}
\multicolumn{1}{|l|}{}                                                                &\multicolumn{1}{l|}{\multirow{6}{*}{\rotatebox[origin=c]{90}{\textbf{Tanh}}}} &      
                                                                                          \multicolumn{2}{l|}{tanh             }                &  2.234 & 1.000 & 37.14  & 1.000      &                   \\ \cdashline{3-9}[6pt/3pt]  
\multicolumn{1}{|l|}{}                                                                & & \multicolumn{2}{l|}{tanh\_cont\_4    }                &  2.256 & 1.010 & 30.39  & 0.818      &                   \\ \cdashline{3-9}[6pt/3pt]  
\multicolumn{1}{|l|}{}                                                                & & \multicolumn{2}{l|}{tanh\_taylor\_9  }                &  2.237 & 1.001 & 35.73  & 0.962      &                   \\ \cdashline{3-9}[6pt/3pt]  
\multicolumn{1}{|l|}{}                                                                & & \multicolumn{2}{l|}{\textbf{tanh\_pade\_4\_4} }       &  2.242 & 1.004 & 32.48  & 0.875      & \textbf{Safe}     \\ \cdashline{3-9}[6pt/3pt]  
\multicolumn{1}{|l|}{}                                                                & & \multicolumn{2}{l|}{\textbf{serp}       }             &  2.147 & 0.961 & 28.36  & 0.770      & \textbf{Ranged}   \\ \cdashline{3-9}[6pt/3pt]  
\multicolumn{1}{|l|}{}                                                                & & \multicolumn{2}{l|}{serp\_clamp  }                    &  2.151 & 0.963 & 31.36  & 0.845      &                   \\ \thickhline

\multicolumn{1}{|l|}{\multirow{17}{*}{\rotatebox[origin=c]{90}{\textbf{CharRNN}}}}    & \multicolumn{1}{l|}{\multirow{17}{*}{\rotatebox[origin=c]{90}{\textbf{LSTM}}}} &
                                                                                          ReLU   & ReLU                                     & NaN     & -     & -      & -                          &                      \\ \cdashline{3-9}[6pt/3pt] 
\multicolumn{1}{|l|}{}                                                                & & sigm & tanh                                       & 79.16   & 1.000 & 1502.93    & 1.000             &                      \\ \cdashline{3-9}[6pt/3pt] 
\multicolumn{1}{|l|}{}                                                                & & sigm\_fastexp\_2& tanh                            & 82.30   & 1.040 & 1406.603   & 0.936             &                      \\ \cdashline{3-9}[6pt/3pt] 
\multicolumn{1}{|l|}{}                                                                & & sigm\_fastexp\_512& tanh                          & 77.65   & 0.981 & 1401.893   & 0.933          &                      \\ \cdashline{3-9}[6pt/3pt] 
\multicolumn{1}{|l|}{}                                                                & & sigm\_taylor\_9& tanh                             & $\infty$& $\infty$ & -       & -                      &                   \\ \cdashline{3-9}[6pt/3pt] 
\multicolumn{1}{|l|}{}                                                                & & sigm\_pade\_4\_4& tanh                            & 78.01   & 0.985 & 1462.484   & 0.973           &                      \\ \cdashline{3-9}[6pt/3pt] 
\multicolumn{1}{|l|}{}                                                                & & sigm          & tanh\_cont\_4                     & 78.68   & 0.994 & 1361.133   & 0.906           &                      \\ \cdashline{3-9}[6pt/3pt] 
\multicolumn{1}{|l|}{}                                                                & & sigm          & tanh\_taylor\_9                   & $\infty$& $\infty$ & -       & -                      &                   \\ \cdashline{3-9}[6pt/3pt] 
\multicolumn{1}{|l|}{}                                                                & & sigm          & tanh\_pade\_4\_4                  & 77.99   & 0.985 & 1407.648   & 0.937                  &                      \\ \cdashline{3-9}[6pt/3pt]
\multicolumn{1}{|l|}{}                                                                & & sigm          & serp                              & 78.29   & 0.989 & 1303.92    & 0.868            &                      \\ \cdashline{3-9}[6pt/3pt] 
\multicolumn{1}{|l|}{}                                                                & & sigm          & serp\_clamp                       & 78.46   & 0.991 & 1367.542   & 0.910                  &                      \\ \cdashline{3-9}[6pt/3pt] 
\multicolumn{1}{|l|}{}                                                                & & \textbf{sigm\_fastexp\_2}& \textbf{serp}          & 79.82   & 1.008 & 1332.127   & 0.886                                 & \textbf{Ranged}      \\ \cdashline{3-9}[6pt/3pt]
\multicolumn{1}{|l|}{}                                                                & & sigm\_fastexp\_2 & serp\_clamp                    & 82.81   & 1.046 & 1184.179   & 0.788           &                      \\ \cdashline{3-9}[6pt/3pt]
\multicolumn{1}{|l|}{}                                                                & & sigm\_fastexp\_512& tanh\_cont\_4                 & $\infty$& $\infty$& -        & -                      &                      \\ \cdashline{3-9}[6pt/3pt]
\multicolumn{1}{|l|}{}                                                                & & sigm\_fastexp\_512& tanh\_pade\_4\_4              & 79.34   & 1.002 & 1446.656   & 0.963                &                      \\ \cdashline{3-9}[6pt/3pt]
\multicolumn{1}{|l|}{}                                                                & & sigm\_fastexp\_512& serp                          & 77.76   & 0.982 & 1150.14    & 0.765                  &                      \\ \cdashline{3-9}[6pt/3pt]
\multicolumn{1}{|l|}{}                                                                & & \textbf{sigm\_fastexp\_512}& \textbf{serp\_clamp} & 79.24   & 1.001 & 1155.745   & 0.769           & \textbf{Safe}        \\ \cdashline{3-9}[6pt/3pt]
\multicolumn{1}{|l|}{}                                                                & & ultra\_fast\_sigmoid$\dagger$& tanh               & 78.88   & 0.996 & 1450.332   & 0.965             &                      \\ \cdashline{3-9}[6pt/3pt] 
\multicolumn{1}{|l|}{}                                                                & & word2vec$\dagger$            & tanh               & 100.7   & 1.271 & 1561.443   & 1.039             &                      \\ \thickhline
                
\end{tabular}
\caption{Performance results for the range of approximations considered. The Rel. columns indicate performance relative to the replaced function (smaller values are better). ($\dagger$) We discuss the performance of \approxfn{ultra\_fast\_sigmoid} and \approxfn{word2vec} later.}
\label{tab:SelectedResults}
\end{table}

We used each approximation in turn to train our three networks for a fixed number of epochs recording the loss and the total time taken. We then compared these values to the non-approximated activation functions to compute the relative loss and relative time taken (Table \ref{tab:SelectedResults}). Relative values are with respect to the function being replaced (rather than \relu{}) and smaller relative values indicate better performance. We used an Intel Xeon E5-2673 v3 @ 2.40GHz (14GB RAM) for the MNIST workloads and a dual-core Intel Xeon E5-2673 v4 @ 2.30GHz (8GB RAM) for the RNN workload. We ran our benchmarks on Azure cloud machines and we provide virtual machine images\footnote{\url{https://drive.google.com/file/d/1trqpemv9BScwt88Xd69zpZKGWM2RMXs3/view?usp=sharing}} for Azure which replicate our results.

In most cases our replacement activation functions either converged with similar loss to the original function or failed to converge entirely (marked $\infty$ in the table). We found a few instances (such as \approxfn{sigm_fastexp_2} in Convnet) for which the loss was drastically lower (44\%). We highlight this case as another example of the difficulty of making definitive statements about the correct choice of activation function. In all cases where the training loss converged our approximations resulted in a reduced overall training time.

We found that two functions (\approxfn{sigm_fastexp_512} and \approxfn{tanh_pade_4_4}) produced the best reduction in training time (\safesmallestsavingpercent{}--\safebiggestsavingpercent{}) whilst working in all cases. We mark these as our chosen \emph{safe} approximations. We also found functions (such as \approxfn{sigm_fastexp_2} and \approxfn{serp}) which produced even better reduction in training times (\rangedsmallestsavingpercent{}--\rangedbiggestsavingpercent{}) but have such significant divergence from the target functions that we cannot argue that they are a suitable replacement in all cases. We mark these as \emph{ranged} approximations suitable for networks (such as these) where the activation input values are roughly within the range of $-5$ to $5$.

We also found performance improvements when using approximations for inference rather than training. For the Character-based RNN (LSTM) model our safe approximations offered \textasciitilde 8\% savings where as our ranged approximations allowed for savings of up to 20\% when performing 1000 sequential inferences. This is potentially more signficant than the reduction in training times since inference is often performed on restricted hardware such as mobile devices.


\section{Comparison with \approxfn{ultra\_fast\_sigmoid}}
\approxfn{ultra_fast_sigmoid} is a fast \sigmoid{} implementation in popular machine learning framework library Theano~\cite{bergstra2010theano}. It is implemented as a piece-wise approximation. 
It is one of only a few approximations available openly as standard in popular machine learning libraries.

\begin{figure}
\centering
\begin{minipage}{.45\textwidth}
  \centering
  \includegraphics[width=0.9\textwidth]{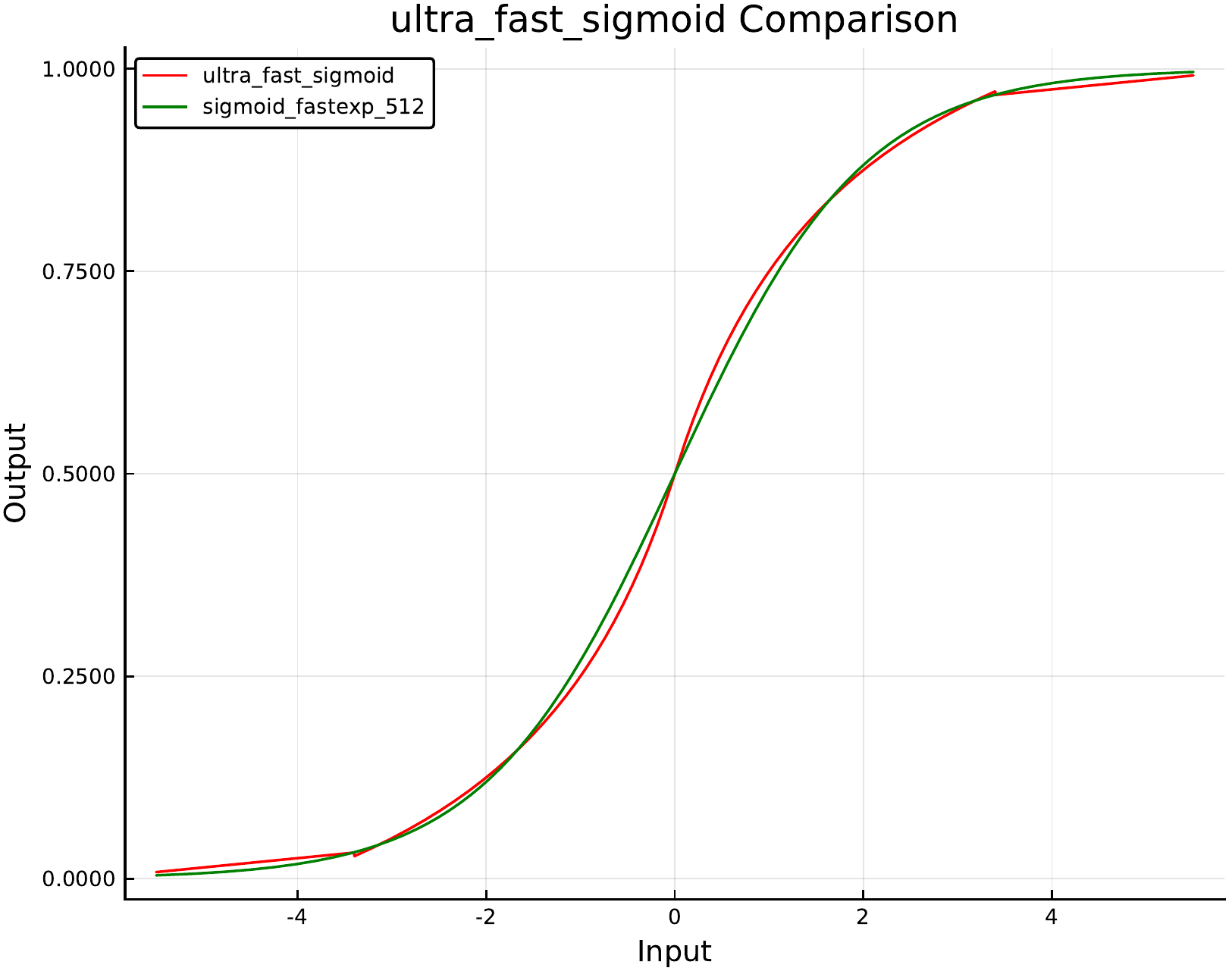}
\end{minipage}%
\begin{minipage}{.45\textwidth}
  \centering
  \includegraphics[width=0.9\textwidth]{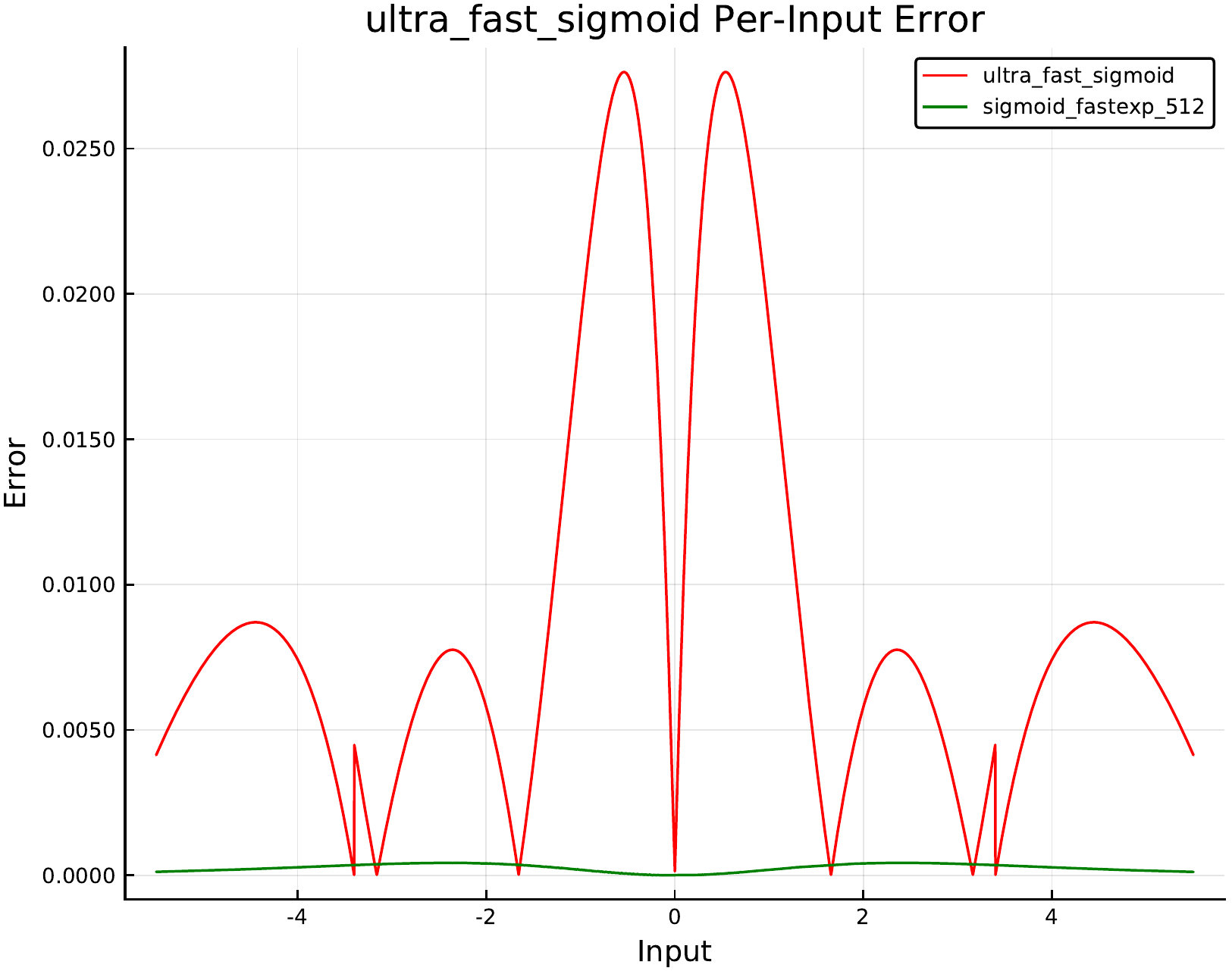}
\end{minipage}
\caption{Shape and relative error of Theano's \approxfn{ultra\_fast\_sigmoid} compared to \approxfn{sigm\_fastexp\_512}.}
\label{fig:TheanoError}
\end{figure}


We compared \approxfn{ultra_fast_sigmoid} to \approxfn{sigm_fastexp_512}. Figure \ref{fig:TheanoError} shows the much greater approximation error for \approxfn{ultra_fast_sigmoid}. For the Autoencoder and CharRNN workloads \approxfn{sigm_fastexp_512} results in lower loss in less time. For the Convnet workload \approxfn{sigm_fastexp_512} is slightly slower but with drastically lower loss. 


\section{Comparison with Word2Vec lookup table}

\begin{figure}
\centering
\begin{minipage}{.48\textwidth}
  \centering
  \includegraphics[width=0.9\textwidth]{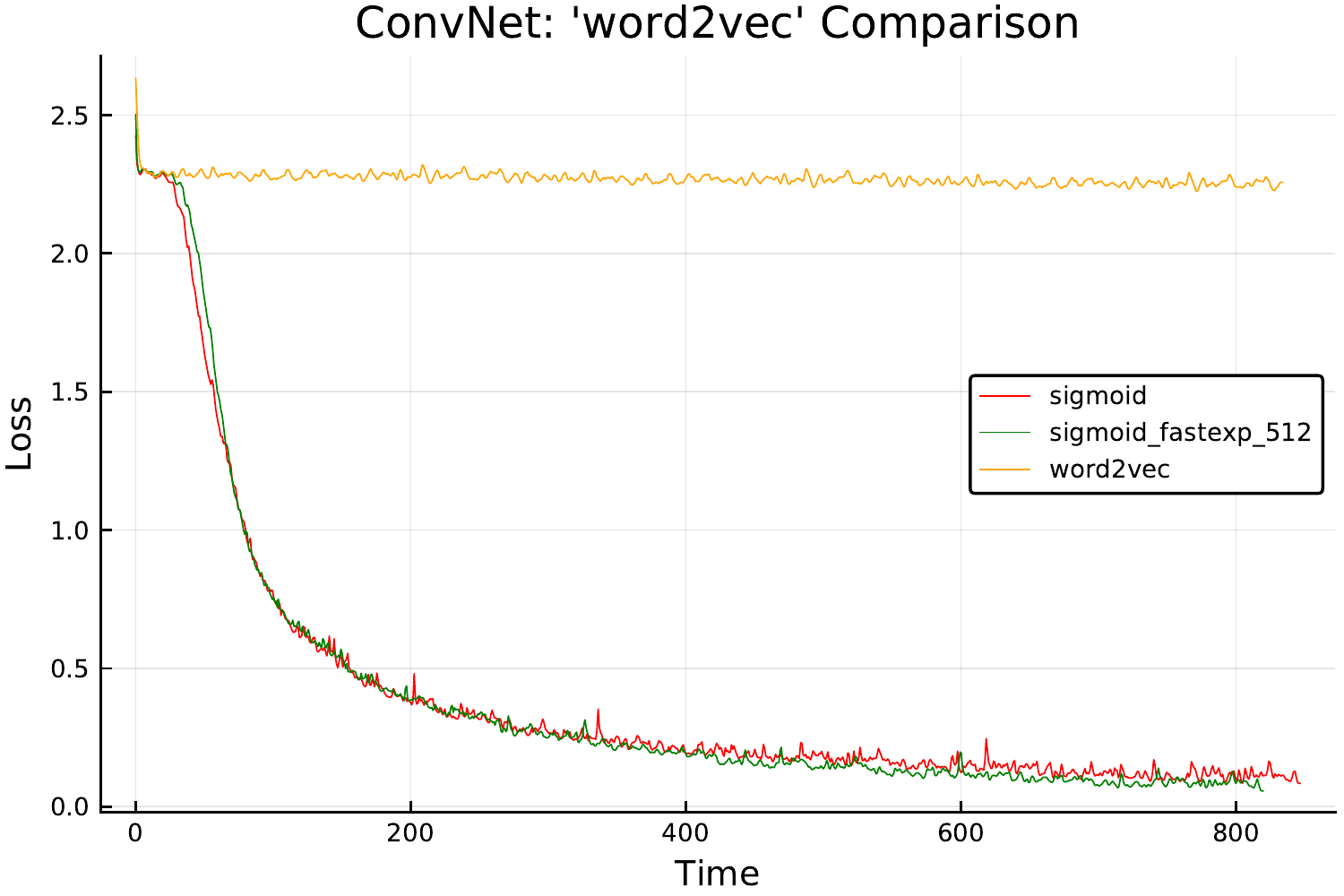}
\end{minipage}
\begin{minipage}{.48\textwidth}
  \centering
  \includegraphics[width=0.9\textwidth]{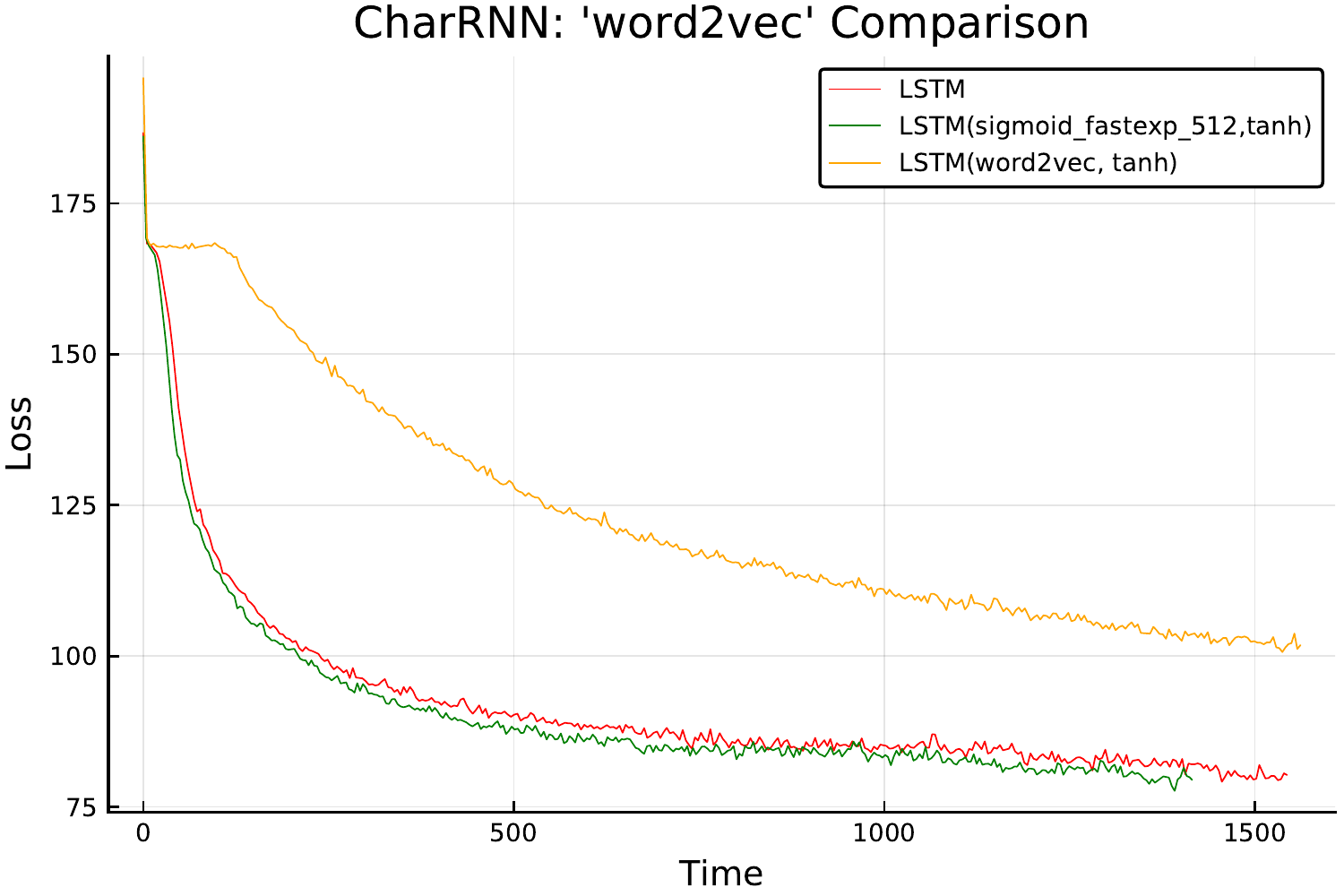}
\end{minipage}
\caption{Loss over time for the \approxfn{word2vec} table based sigmoid function. }
\label{fig:LossResultsTable}
\end{figure}

Word2Vec makes use of shallow 2-layer networks to create word embeddings. In these models the authors used an approximated sigmoid implementation which makes use of a 1000 element lookup table. We have compared this lookup table to our approximations in our test models.

Our results (Table \ref{tab:SelectedResults}) show that applying this approximation results in more loss for a similar or slightly reduced training time. When looking at the loss over time (Figure~\ref{fig:LossResultsTable}) we see that in some cases training fails to make progress. We believe this occurs due to the lack of interpolation in the table lookup resulting in quantisation of outputs. As a result small incremental changes to the weight values may not result in a change of output. This could potentially be migitated with use of a different optimiser or by adding interpolation (at the cost of a performance reduction).

Again, \approxfn{sigmoid_fastexp_512} results in less loss in less training time.

\section{Approximations in TensorFlow}

TensorFlow is one of the most commonly used machine learning libraries. Despite the fact that it is a Python library it achieves high performance through native implementation of the core functions. As such this makes experimentation with novel activation functions difficult: one must inject a low-level implementation and then provide a mechanism to reference it from the high-level code. Flux does not suffer from this issue because the entire system is implemented in Julia (a relatively high performance language) and alternative activation functions can be straightforwardly applied on a level playing field with the default options.

Despite being unable to directly evaluate our new activation functions in TensorFlow we were able to identify the optimisation space by measuring the performance of the simplest possible activation function, the identity function.





\begin{figure}
\centering
  \centering
  \includegraphics[width=0.4\textwidth]{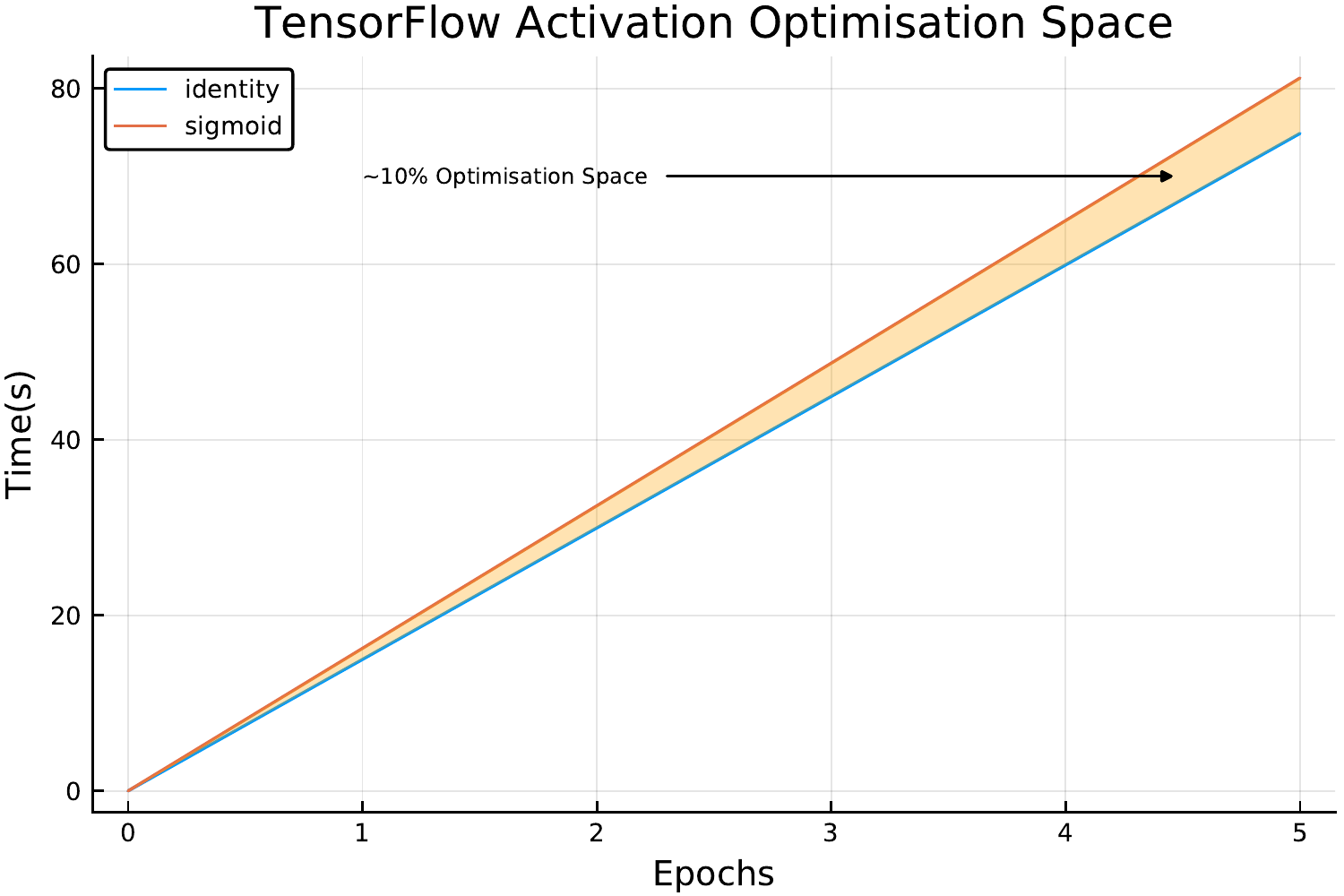}
  \includegraphics[width=0.4\textwidth]{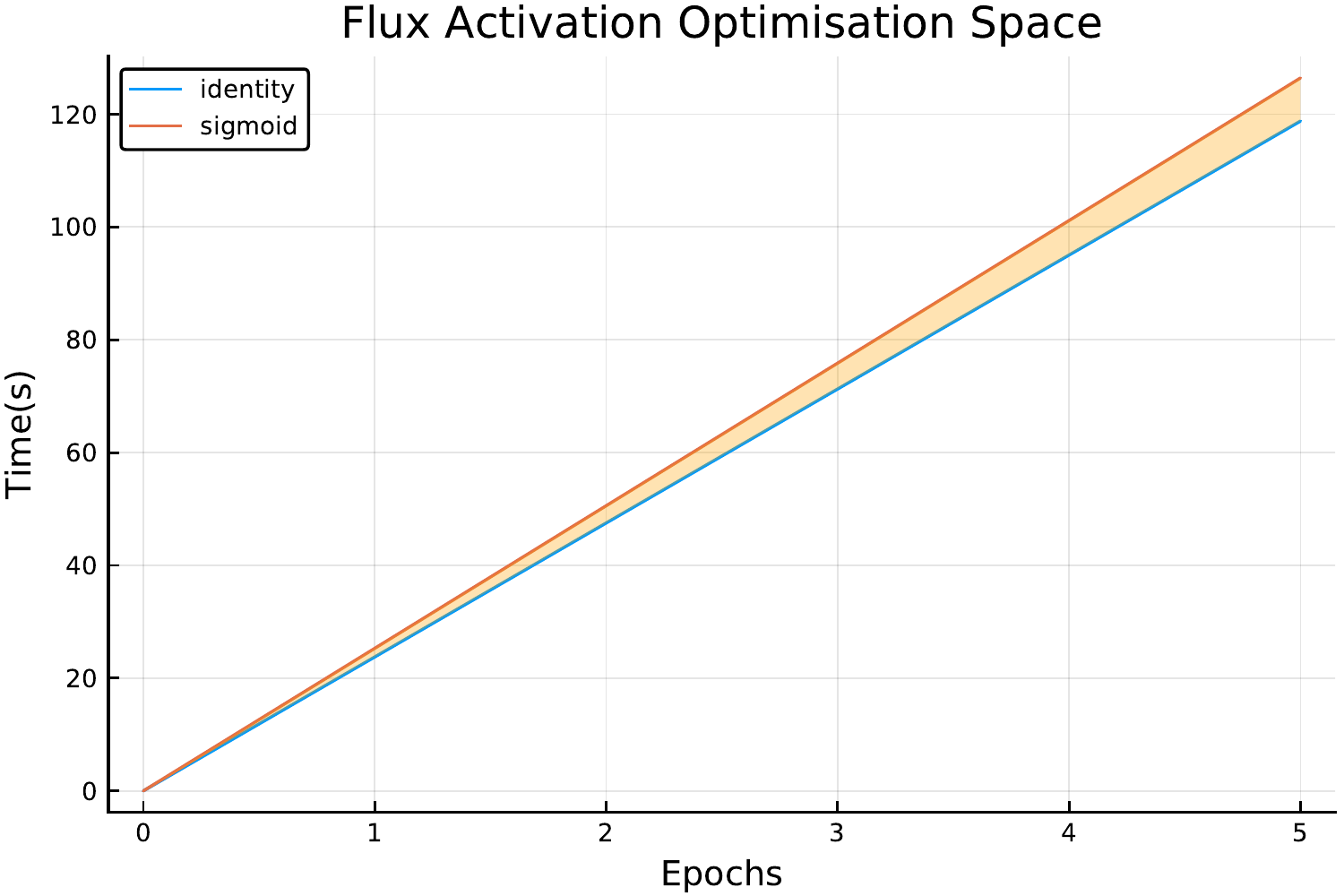}
  \captionof{figure}{Training time per epoch for the MNIST classifier in TensorFlow and Flux using \sigmoid{} and \identity{} activation functions.}
  \label{fig:TensorFlowEpochGraph}
\end{figure}

Figure \ref{fig:TensorFlowEpochGraph} shows the time taken to train a simple MNIST classifier (one convolution layer and two dense layers) when using \sigmoid{} and \identity{} in both TensorFlow (left) and Flux (right). The approximations we have discussed in this paper fall within the region between these two lines. Even with this simple model this demonstrates that there is scope for improving model performance in TensorFlow by optimising activation functions.

\section{Threats to validity}

Although we have tried to evaluate our activation functions on three representative workloads it is not possible to say for sure how well they will work in the general case. However, the relative errors in our safe approximations are so small that we would expect them to be drop-in replacements. 

Neural networks are commonly trained offline on large compute clusters whereas inference happens with interactive latencies and increasingly on limited hardware (such as mobile phone handsets). The majority of our results focus on training times because training loss provides a convenient measurement to check that the activation function is performing well. However, we also found that our approximations improve inference and mention this in Section \ref{sec:PerformanceResults}.

Our results only consider the performance on CPUs whereas much training (and inference) happens on GPUs or specialist hardware such as TPUs. We therefore cannot say how our approximations perform in these circumstances. We argue that our safe approximations are useful even if only applied to CPUs since they generally reduce training times with no impact on loss. It would be particularly interesting to consider hardware implementations of these approximations for specialist hardware. We leave this for future work.

\section{Conclusion}

We have shown that approximation of activation functions in neural networks can improve the training and inference time without a negative effect on the accuracy of the network.

We investigated a range of functions and propose two safe approximations \approxfn{sigm_fastexp_512} and \approxfn{tanh_pade_4_4} for \sigmoid{} and \tanh{} respectively. These approximations produce faster training and inference times without damaging prediction accuracy. Our ranged approximations produced even larger speedups but will not work for all networks.

As such we think these functions are candidates for inclusion as standard options in machine learning libraries. We also showed that these functions outperform existing approximations in these libraries.

In the future we hope to expand on this work by integrating approximation as a standard option into many machine learning libraries so that it can be used to improve training time on a larger scale. Additionally we wish to analyse the effect of approximations on large and complex networks and hardware to understand if there is a structure which may cause approximations to not be beneficial.

\bibliographystyle{unsrt}
\bibliography{biblio}

\end{document}